\documentclass[letterpaper, 10 pt, conference]{ieeeconf}  

\IEEEoverridecommandlockouts                              

\usepackage{graphics} 
\usepackage{epsfig} 
\usepackage{times} 
\usepackage{amsmath} 
\usepackage{amssymb}  
\usepackage[tight,footnotesize]{subfigure}
\usepackage{enumerate}
\usepackage{cite}

\title{\LARGE \bf
SeqLPD: Sequence Matching Enhanced Loop-Closure Detection Based on Large-Scale Point Cloud Description for Self-Driving Vehicles
}

\author{Zhe~Liu$^{1*}$, Chuanzhe~Suo$^{1*}$, Shunbo~Zhou$^{1}$, Huanshu~Wei$^{1}$, Yingtian~Liu$^{1}$, Hesheng~Wang$^{2}$, Yun-Hui~Liu$^{1}$\\$^{1}$The Chinese University of Hong Kong. $^{2}$Shanghai Jiao Tong University.
}
\begin{document}

\maketitle
\thispagestyle{empty}
\pagestyle{empty}

\begin{abstract}
Place recognition and loop-closure detection are main challenges in the localization, mapping and navigation tasks of self-driving vehicles. 
In this paper, we solve the loop-closure detection problem by incorporating the deep-learning based point cloud description method and the coarse-to-fine sequence matching strategy.
More specifically, we propose a deep neural network to extract a global descriptor from the original large-scale 3D point cloud, then based on which, a typical place analysis approach is presented to investigate the feature space distribution of the global descriptors and select several super keyframes.
Finally, a coarse-to-fine strategy, which includes a super keyframe based coarse matching stage and a local sequence matching stage, is presented to ensure the loop-closure detection accuracy and real-time performance simultaneously.
Thanks to the sequence matching operation, the proposed approach obtains an improvement against the existing deep-learning based methods.
Experiment results on a self-driving vehicle validate the effectiveness of the proposed loop-closure detection algorithm.

\end{abstract}

\section{Introduction}

Autonomous navigation is paramount significant in robotic community such as helping self-driving vehicles and unmanned aerial vehicles achieve full autonomy \cite{Li2019}. Place recognition and loop-closure detection, in particular, are crucial challenges of accurate navigation, since they provide candidates for loop-closure, which is essential for drift-free localization and global consistent mapping \cite{iros18kim}.
Current solutions for place recognition and loop-closure detection mainly fall into two categories, image-based and 3D point cloud-based.
Many successful image-based approaches were proposed in the literature, due to the feasibility of extracting visual feature descriptors (e.g. SURF or ORB \cite{orb}). These visual feature descriptors can be aggregated into a global descriptor of the scene for place recognition \cite{visualpr}. Unfortunately, this kind of solutions are unreliable due to its non-robustness under different season or weather conditions, and under different viewpoints \cite{iros18yin}.
3D point cloud-based solutions, on the other hand, does not suffer from changes in external illumination, and does not suffer as much as vision solutions when changes in viewpoint are present. This paper therefore considers 3D point cloud for their potential to provide robust loop-closure detection in large-scale environments.

\begin{figure}[!t]
\centering
\includegraphics[width=0.95\columnwidth]{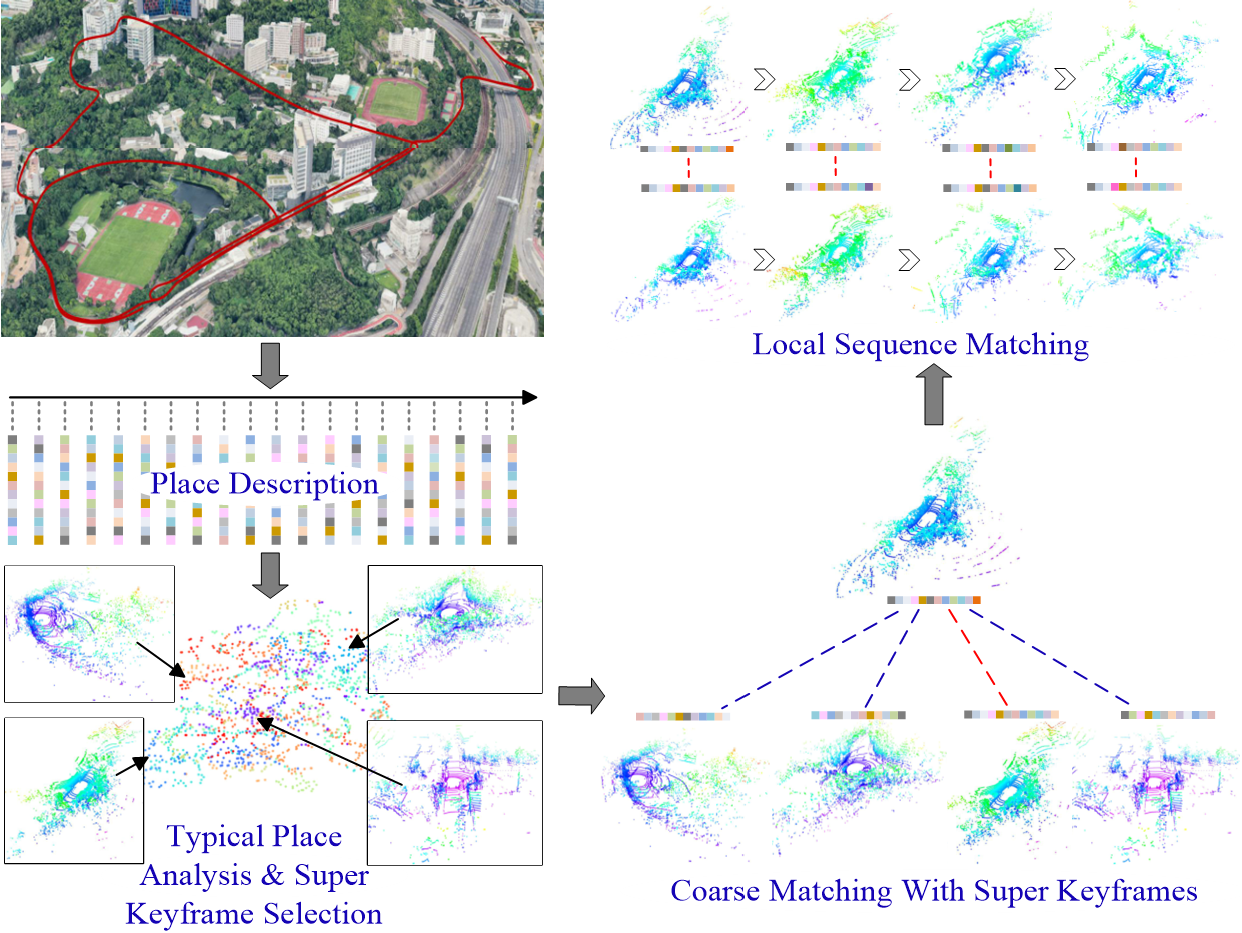}
\caption{System framework of the proposed SeqLPD approach.}
\label{figsys}
\vspace{-0.5cm}
\end{figure}

%
%
%
%

In \cite{myiccv}, we proposed a large-scale place description network (LPD-Net) and achieved the state-of-the-art place retrieval performance in the Oxford RoboCar Dataset \cite{pointnetvlad}. However, directly implementing LPD-Net can not fulfill the accuracy and real-time requirements in loop-closure detection tasks of self-driving vehicles. Compared with dataset testings, the environments in practical applications are more complex and the sensor information mostly contains lots of noises and uncertainties, especially in some industrial and field applications. What's more, the heavy computation and memory requirements of the complicated network also limit its applications in practical vehicle platforms.

In order to improve the loop-closure detection accuracy in practical implementations and ensure the real-time requirement simultaneously, we present SeqLPD (as shown in Fig. \ref{figsys}) in this paper, which incorporates the point cloud description network and the coarse-to-fine sequence matching strategy. The contributions of our work can be summarized as follows: Firstly, a lightweight place description network was adopted to extract a discriminative and generalizable global descriptor from the original large-scale 3D point cloud. Based on which, a typical place analysis approach is presented to select super keyframes. Then a coarse-to-fine sequence matching approach is developed to improve the loop-closure detection accuracy with a feasible online searching time. Finally, the proposed approach is evaluated in large-scale environments with dynamic objects and slope terrains, which validate the accuracy, robustness and the real-time performance in practical applications.

\section{Related Work}



We identify the current 3D point cloud based loop-closure detection solutions into three main trends: approaches based on handcrafted local and global features, based on planes or objects, and based on 3D point cloud learning.


Handcrafted local features, such as SHOT \cite{shot} and FPFH \cite{fpfh}, are usually based on pre-defined keypoints and are not suitable for large-scale place recognition tasks in outdoor dynamic scenarios, while the global descriptors, such as M2DP \cite{m2dp} and ESF \cite{esf}, are usually tailored to specific tasks and have poor generalization abilities. Taking into account the limitations of the above descriptors, 
a plane-based place recognition method was proposed in \cite{plane}, the covariance of the plane parameters was adopted for matching. The drawback of this approach is that it only applies to small and indoor environments, and the plane model assumption is not always valid in some practical scenes.
SegMatch \cite{segmatch} and SegMap \cite{segmap} presented a place matching method based on local segment descriptions. However, they require that there are enough static objects in the environment and need to build a dense local map by accumulating a stream of original point clouds to solve the local sparsity problem.


Recently, deep neural network was introduced for 3D point cloud feature learning and achieved state-of-the-art performance. Some work attempts to convert point cloud input to a regular 3D volume representation to alleviate the orderless problem of the point cloud, such as the 
3D ShapeNets \cite{shapenet}. 
On the other hand, different from volumetric representation, Multiview CNNs \cite{multiviewcnn} projects the 3D data into 2D images so that 2D CNN can be performed.
By achieving the permutation invariance in the network, PointNet \cite{pointnet} makes it possible to learn features directly from the raw point cloud data. Although PointNet has achieved superior performance on small-scale shape classification and recognition tasks, it did not scale well for large-scale place recognition problem.
PointNetVLAD \cite{pointnetvlad} is the first work that directly applies 3D point cloud to large-scale place recognition, but this method does not consider local feature extraction adequately, and the spatial distribution information of local features has also not been considered.

\section{System Framework}

As shown in Fig. \ref{figsys}, the proposed SeqLPD has three parts:

\begin{itemize}
\item \textbf{Large-scale place description}: The original 3D laser point cloud is used as system input directly. For each point $p_i$ in the point cloud, 
    both the local features and the transformed coordinates are utilized as the network input. Then we present a lightweight deep neural network to generate a global descriptor to uniquely describe the input point cloud.
\item \textbf{Place clustering and typical place analysis}: The generated global descriptors are stored with corresponding position information in order to generate a place descriptor map. Then we analyze the feature space distribution characteristics of the global descriptors to generate several descriptor clusters and select one typical place in each cluster. The descriptors corresponded to these typical places are defined as the super keyframes.
\item \textbf{Coarse-to-fine sequence matching}: A coarse-to-fine strategy is proposed for loop-closure detection. In the coarse matching stage, the global descriptor of the new input point cloud is compared with all the super keyframes firstly to find out the matched cluster. Then in the fine matching stage, local sequence matching strategy is utilized around each place in the matched cluster to find out the accurate location of the input point cloud, thus achieving the loop-closure detection.
\end{itemize}

The presented coarse-to-fine matching strategy contributes to reducing the computational and storage complexity. Based on the proposed super keyframe selection method and the local sequence matching strategy, both the accuracy and the real-time performance of the loop closure detection in large-scale complex environments can be guaranteed.

\section{Large-Scale Place Description}

\begin{figure}[!h]
\centering
\includegraphics[width=1\columnwidth]{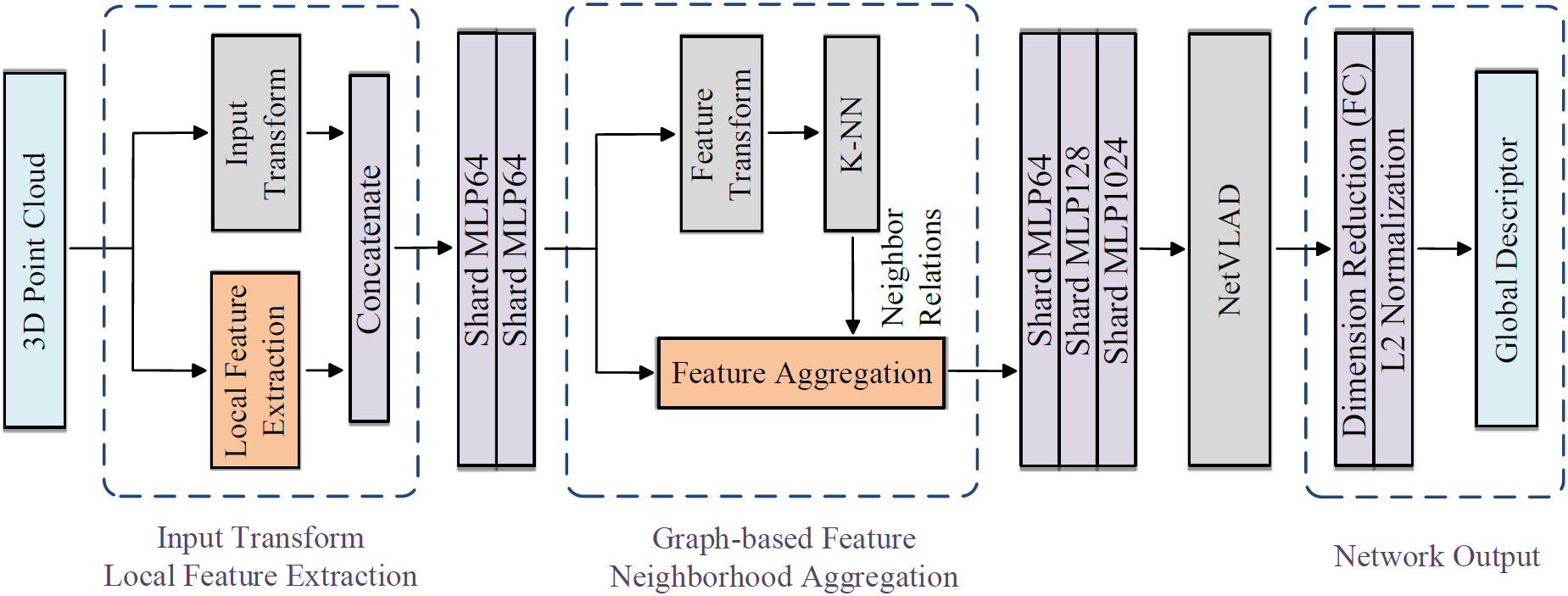}
\caption{Network Architecture.}
\label{fignet}
\end{figure}

The proposed network is shown in Fig. \ref{fignet}. This is a lightweight variant of our LPD-Net \cite{myiccv} and can be implemented on practical vehicle platforms with limited computing and storage resources. In order to avoid repetitions and stress the key points, here we only briefly describe the whole network structure and present the most important modules.

Compared with the existing point cloud description network \cite{pointnet,pointnetvlad}, we present two new modules to improve the place recognition accuracy, i.e., local feature extraction and graph-based feature neighborhood aggregation. The former one aims to extract local features around each point, while the later one aims to reveal the spatial distributions of the similar local structures. We believe that this information is of great importance in SLAM applications and may greatly affect the performance in large-scale environments. Then similar with \cite{pointnetvlad}, the NetVLAD is introduced into the proposed network to generate the final global descriptor (in the form of a 256-dimensional vector) 
and the lazy quadruplet loss is utilized as the network loss function. 

\subsection{Local Feature Extraction}
Existing researches have validated that local features are effective in 3D outdoor scene interpretation tasks and large-scale localization tasks for self-driving applications \cite{iros18kim,segmatch,localfeature,baiducvpr}. So in this paper, $k$ nearest neighboring points are considered to describe the local spatial structure around each point and two types of local features are calculated:
\begin{figure}[!h]
\centering
\includegraphics[width=0.95\columnwidth]{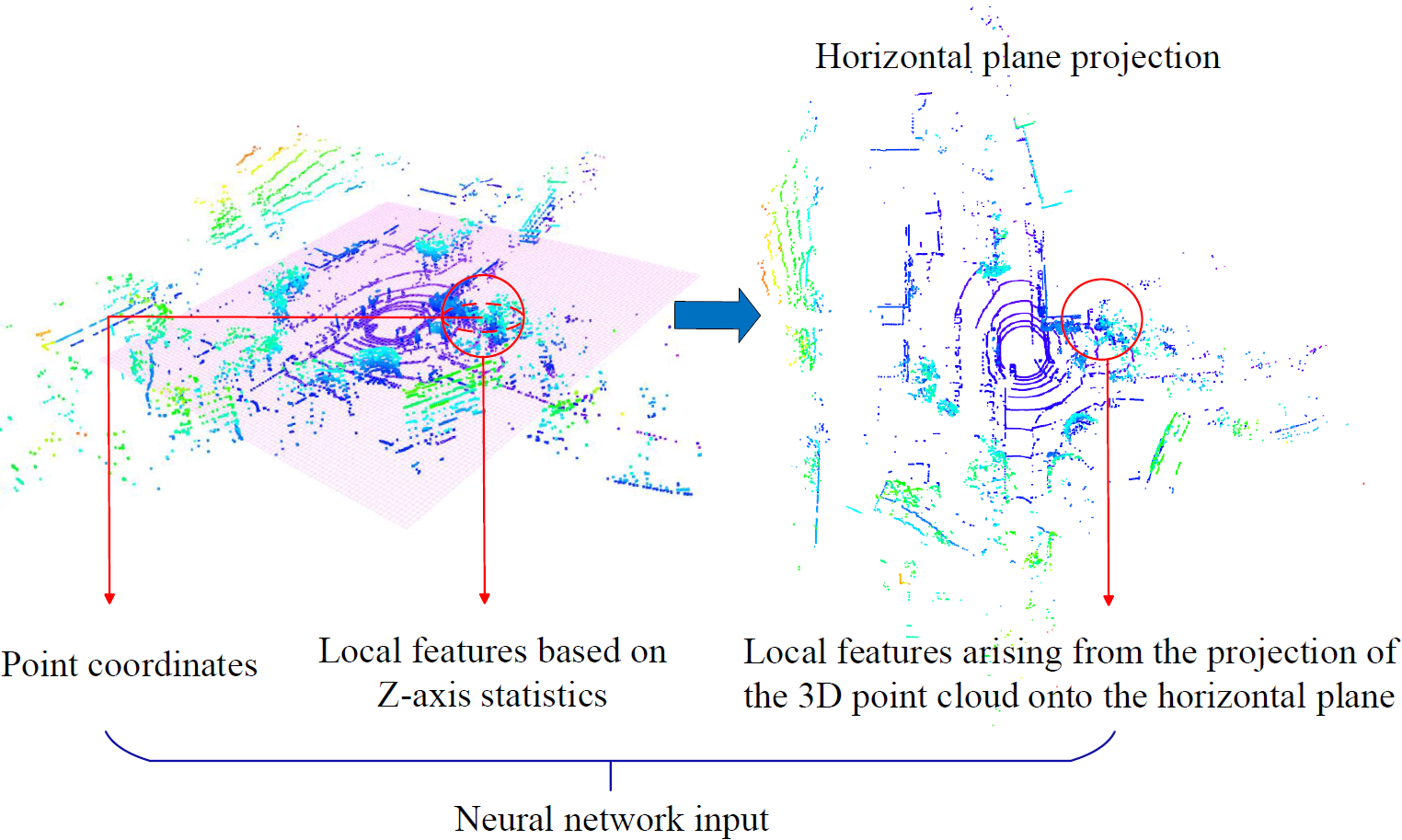}
\caption{Local feature extraction and network inputs.}
\label{figfeature}
\end{figure}

\noindent {\bf Local features based on Z-axis statistics}: the maximum height difference $\Delta Z_{i,max}$ of all the $k$ neighbors around point $i$ and their height variance $\sigma Z_{i,var}$.

\noindent {\bf Local features arising from the projection of the 3D point cloud onto the horizontal plane}: 2D scattering $S^{2D}_{i}=\lambda^{2D}_{i,1}+\lambda^{2D}_{i,2}$ and linearity $\frac{\lambda^{2D}_{i,2}}{\lambda^{2D}_{i,1}}$. $\lambda^{2D}_{i,1}$ and $\lambda^{2D}_{i,2}$ are eigenvalues of the 2D covariance matrix calculated by projecting the neighboring points onto the 2D horizontal plane.


As shown in Fig. \ref{fignet} and \ref{figfeature}, we also consider the original coordinates $(x_i,y_i,z_i)$ of each point $i$ as network input, but in order to unify the viewpoint, the coordinates are transformed by an input Transformation Net \cite{pointnet} to ensure the rotational translation invariance. 

\subsection{Graph-Based Feature Neighborhood Aggregation}

\begin{figure}[!h]
\centering
\includegraphics[width=0.9\columnwidth]{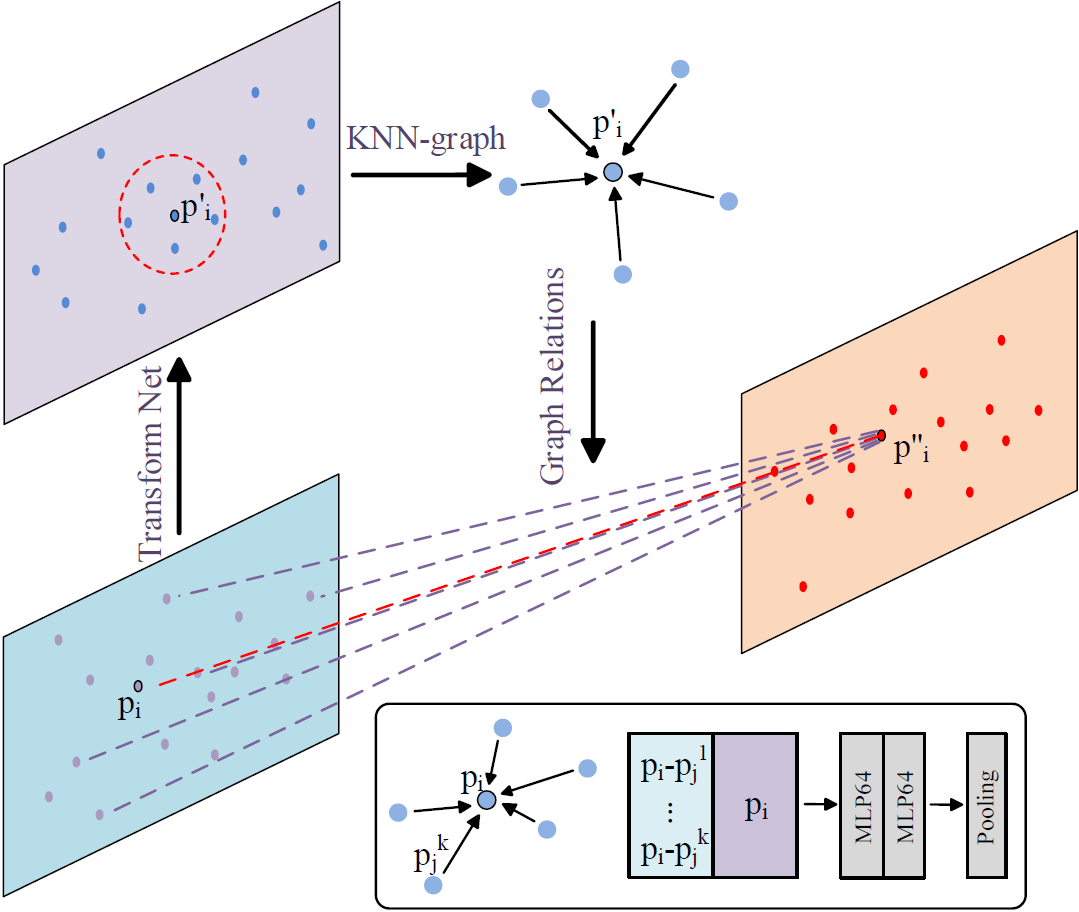}
\caption{Graph-based feature aggregation. $p_i$ represents the feature vector of point $i$, point $j^k$ represents the $k^{th}$ feature space neighbor of point $i$.}
\label{figgraph}
\vspace{-0.4cm}
\end{figure}

A large-scale point cloud mostly consists of rich 3D structures of the surrounding environments and their spatial distribution relationships, such as the relative orientation between two buildings with cube point cloud shapes, or the relative distance between two trees with point cloud clusters. Similar local point cloud structures in different locations usually have similar local features, which can be utilized as a main judgment for place recognition.
We introduce the Graph Neural Network to investigate the intrinsic relationships between each composition in the point cloud. 
More specifically, we first use the Transformation Net on point features and calculate a transformation matrix in the feature space, then the point features are transformed into a unified viewpoint and thus the rigid motion invariance can be guaranteed. As shown in Fig. \ref{figgraph}, a KNN aggregation is implemented on each point to find $k$ nearest neighbors in the new feature space. These feature space relations are further utilized to build the graph in the previous feature space and then we aggregate the neighbor features into each point. Finally, two MLP networks are used to update feature space neighbor relations and a max pooling operation is implemented to aggregate $k$ edge information of each point into a vector. 

In the proposed network, similar local point cloud structures in different locations will be aggregated in the feature space since they usually have similar local features. What's more, since the edge relation is defined as the combination of $p_i-p_j^k$ and $p_i$, the original coordinate information will also be considered in the feature vector $p_i$, i.e., the local spatial distributions and their relative relationships can also be considered in the feature aggregation. So we can reveal the spatial distributions of the similar local structures in the input point cloud adequately. This implies that the proposed network can learn the static geometric structure information in the whole point cloud, thus improving the robustness to dynamic objects and local sparsity problems.


\section{Loop-Closure Detection}
In this section, we first investigate the feature space distribution characteristics of the descriptors and select out the typical places. Then we utilize a coarse-to-fine matching strategy for loop-closure detection in order to ensure the accuracy and real-time performance simultaneously. Please note that the typical places are selected in the feature space (descriptor distribution space) without considering their Cartesian space locations, but in the local sequence matchings, the Cartesian space location of each place in the sequence and the corresponding location relations are considered in order to achieve an accurate result.

\subsection{Place Clustering}
We investigate the feature space distribution of the global descriptors and generate descriptor clusters. Canopy based method and K-means based method are two classes of promising methods for high dimensional space clustering tasks. In our case, we find that K-means is much better since the clustering performance of Canopy depends largely on the initial cluster centers. So we choose K-means++ \cite{kmeans} in this paper. More specifically, we evaluate the sum of distortions under different cluster number $K$ and utilize the Elbow method to determine the optimal $K$ value. What's more, we introduce an additional constraint which requires that the $L_2$ distance from each global descriptor to its corresponding cluster center is lower than $D$, where $D$ is an environment related parameter which defines the $L_2$ distance threshold of two global descriptors which can be recognized as the similar places.

\subsection{Typical Place Selection and Super Keyframe Generation}
In each cluster, the global descriptor with the nearest $L_2$ distance to the cluster center is selected as the super keyframe (the corresponding place is defined as the typical place) and other global descriptors in this cluster are restored in a descriptor index which corresponds to this super keyframe. Then we can obtain $K$ super keyframes and $K$ global descriptor indices. The selected super keyframes contain all types of the characteristic places in the whole environment and each type of them has at least one super keyframe. 
After generating the super keyframes and their corresponding descriptor clusters, we also build a KD-Tree for each descriptor cluster in order to facilitate the following place matching tasks.
\subsection{Coarse Matching and Local Sequence Matching}

In the coarse matching stage, the global descriptor of the new input point cloud is compared with all the $K$ super keyframes firstly to find out the matched cluster by calculating the $L_2$ distances. Then in the fine matching stage, local sequence matching strategy is utilized around each place in the corresponding global descriptor index of the matched cluster (the cluster corresponding to the matched keyframe) to find out the accurate location of the input point cloud, thus achieving the loop-closure detection task.

In particular, the fine matching algorithm is tailored based on SeqSLAM \cite{seqslam}, a robust visual SLAM framework, to make it suitable for point cloud data. The basic idea of fine matching is that, instead of finding the global best match frame relative to the current frame, we look for the best candidate matching frame within every local sequence. To do this, the fine matching process is divided into two components:

\begin{itemize}
\item Local Best Recognition. Local best recognition towards to find all the frames within local neighborhoods that are the best match for the current frame, which is conducted by calculating the difference between two frames based on the $L_2$ distances of the global descriptors that extracted by our network, and a difference matrix would be generated as shown in Fig. \ref{figexp2}.
\item Sequence Matching. To match the target place sequences, a search is performed through the current difference matrix with a searching window.
At each reference frame, the search projects several trajectories based on different possible velocities. The trajectory velocity is ranged from $V_{min}$ to $V_{max}$. The $L_2$ distance based difference score is calculated in each trajectory line. The trajectory with the minimum score is the best match.
\end{itemize}

\subsection{Discussion}
The environments in practical self-driving applications are mostly complex and contain lots of dynamic objects. The extremely similar places in the environment also may not be avoided, especially in some industrial and field applications (such as in the logistic warehouses and port terminal environments). Furthermore, the original point cloud also contains lots of noises and suffers from the local sparsity and appearance difference problems. Directly utilizing the single frame place matching for loop-closure detection tasks can not ensure the accuracy and robustness performance. To overcome the above challenges, we introduce the sequence matching approach with local geographical verifications to achieve an accurate loop-closure detection result. The coarse-to-fine strategy is designed to accelerate the place recognition process and thus ensuring the real-time performance.

%

\section{Experiments}


As shown in Fig. \ref{fignet}, the proposed deep neural network has three main modules: local feature exaction (LFE), graph-based feature aggregation (GFA) and NetVLAD. In LFE, we select $k=20$ nearest points to generate the local neighborhood of each point. In GFA, the $k$ in KNN aggregation is also set to 20. NetVLAD is the same as that in \cite{pointnetvlad}, where the lazy quadruplet loss parameters are set to $\alpha=0.5, \beta=0.2, P_{pos}=2, P_{neg}=18$. All experiments are conducted on a 1080Ti GPU with TensorFlow.

\subsection{Place Recognition Results in RobotCar Datasets}

We train and evaluate the proposed network on the Oxford RobotCar dataset \cite{oxford}, in which the 3D point cloud submap is made up of point clouds within the car's 20m trajectory. Each submap contains 4096 points with a normalized range of $[-1,1]$. We use 44 sets, 21,711 submaps for training and 3030 submaps for testing. 
The data collection of 44 sets is in different seasons, different time and different weathers, and we query the same scene in these sets for place recognition.
We use Recall to evaluate the ability of place recognition to see if there is a real scene in the top N scenes closest to it. We compare it with Average Recall@1 and Average Recall@1\%.


\vspace{-0.3cm}
\begin{table}[!h]\scriptsize
\renewcommand{\arraystretch}{1.2}
\caption{Comparison results of the place recognition.}
\label{tabexp}
\centering
\begin{tabular}{ccc}
\hline
\hline
 & Ave. Recall @1\% & Ave. Recall @1\\
\hline
PN-VLAD baseline\cite{pointnetvlad} & $81.01$ & $62.76$\\
PN-VLAD refine\cite{pointnetvlad} & $80.71$ & $63.33$\\
LPD-Net\cite{myiccv} & $94.92$ & $86.28$\\
\hline
LPD-Light (our) & ${89.55}$ & ${77.92}$\\
LPD-Light + SequenceMatching (our) & ${95.81}$ & ${87.15}$\\
\hline
\hline
\end{tabular}
\end{table}

\vspace{-0.3cm}
\begin{table}[!h]\scriptsize
\renewcommand{\arraystretch}{1.2}
\caption{Comparison results of the computation and memory required.}
\label{tabexp2}
\centering
\begin{tabular}{cccc}
\hline
\hline
 & Parameters & FLOPs & Runtime\\
\hline
PN-VLAD\cite{pointnetvlad} & $1.978\rm{M}$ & $411\rm{M}$ & $13.09\rm{ms}$\\
LPD-Net\cite{myiccv} & $1.981\rm{M}$ & $749\rm{M}$ & $23.58\rm{ms}$\\
\hline
LPD-Light (our) & $1.980\rm{M}$ & $614\rm{M}$ & $18.88\rm{ms}$\\
\hline
\hline
\end{tabular}
\end{table}
\vspace{-0.3cm}

We compare our approach 
with the state-of-the-art PointNetVLAD (PN-VLAD baseline and PN-VLAD refine) \cite{pointnetvlad} and our previous work LPD-Net \cite{myiccv} to show the performance of the proposed network. 
Comparison results are shown in Tab. \ref{tabexp}, where the LPD-Light represents the lightweight neural network (without sequence matching) presented in Section IV. 
From Tab. \ref{tabexp} we can find that LPD-Light is better than PointNetVLAD, which increases the place recognition accuracy from $81.01\%$ to $89.55\%$, since we introduce the local features and graph-based feature aggregations to reveal the spatial distribution of the input point cloud adequately. 
For single frame place recognition, LPD-Net is slightly better than LPD-Light and achieves the best results. However, from Tab. \ref{tabexp2} we can find that LPD-Net is much heavier than LPD-Light, in the aspects of total network parameter scales (Parameters), required floating-point operations (FLOPs), and network runtime per frame (Runtime). What's more, in LPD-Net, we need about 940ms for local feature extractions, but in LDP-Light we only need 310ms, thus greatly improving the real-time performance. Furthermore, thanks to the additional sequence matching, the accuracy can be improved to $95.81\%$, which is better than LPD-Net.
Please note that in order to facilitate the comparison, in each test, we simply select five consecutive point clouds and evaluate their place recognition accuracy independently. If the retrieval tasks succeed more than three times, we define that a correct matching occurs. This is not a rigorous index for accuracy evaluation, however, we believe that the comparison results in Tab. \ref{tabexp} can validate the effectiveness of the proposed approach.


\subsection{Place Clustering and Matching in KITTI Dataset}

\begin{figure}[!h]
	\centering
	\includegraphics[width=1\columnwidth]{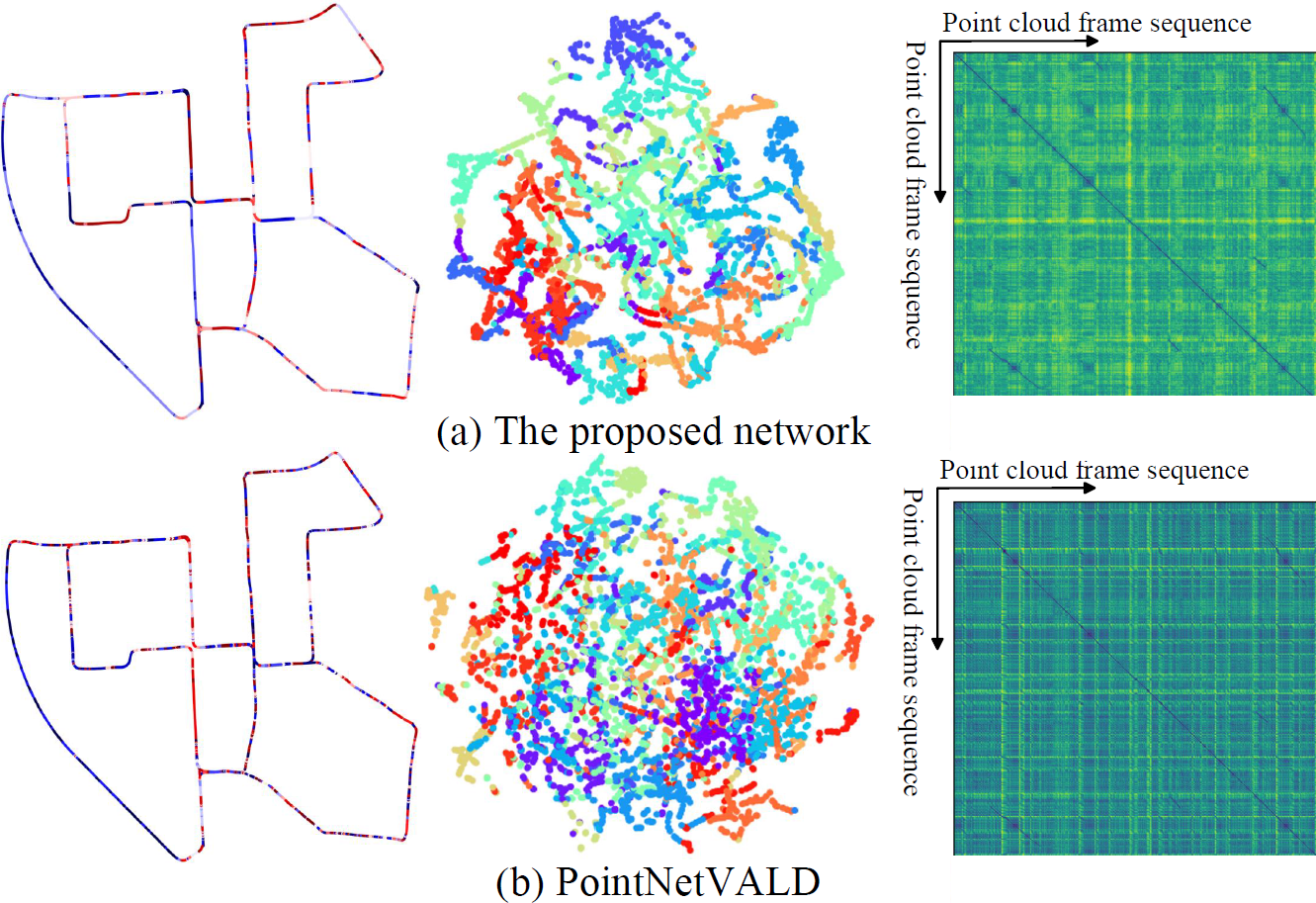}
	\caption{Place clustering and matching results. Middle figures show the t-SNE visualization of the clustering results using global descriptors. The color of each location in left figures represents its belonged clusters. Right figures show the $L_2$ distances between the generated descriptors along the whole point cloud frame sequence, the darker color represents the smaller $L_2$ distance.}
	\label{figclu}
	\vspace{-0.2cm}
\end{figure}

We compare our network with PointNetVLAD in KITTI dataset to show the place clustering and matching performance. As shown in Fig. \ref{figclu}, the clustering results of the proposed network are better than those of the PointNetVLAD, since we achieve a more dense clustering results and the distance between each cluster pairs is more larger than that in PointNetVLAD. What's more, the right figures in Fig. \ref{figclu} show that the descriptors generated by our network are more discriminative than PointNetVLAD.


\subsection{Real Experimental Results}


We use an autonomous tractor as the experiment platform, which is equipped with cameras, millimeter wave radars, IMU and encoders.
In particular, a Velodyne VLP-16 3D laser is equipped that allows for collecting real-time 3D point cloud data of the surrounding environment.
A computer with a single NVIDIA 1080Ti GPU, i7-6700 processor, and 16G RAM is used.



\begin{figure*}[!t]
\centering
\includegraphics[width=2\columnwidth]{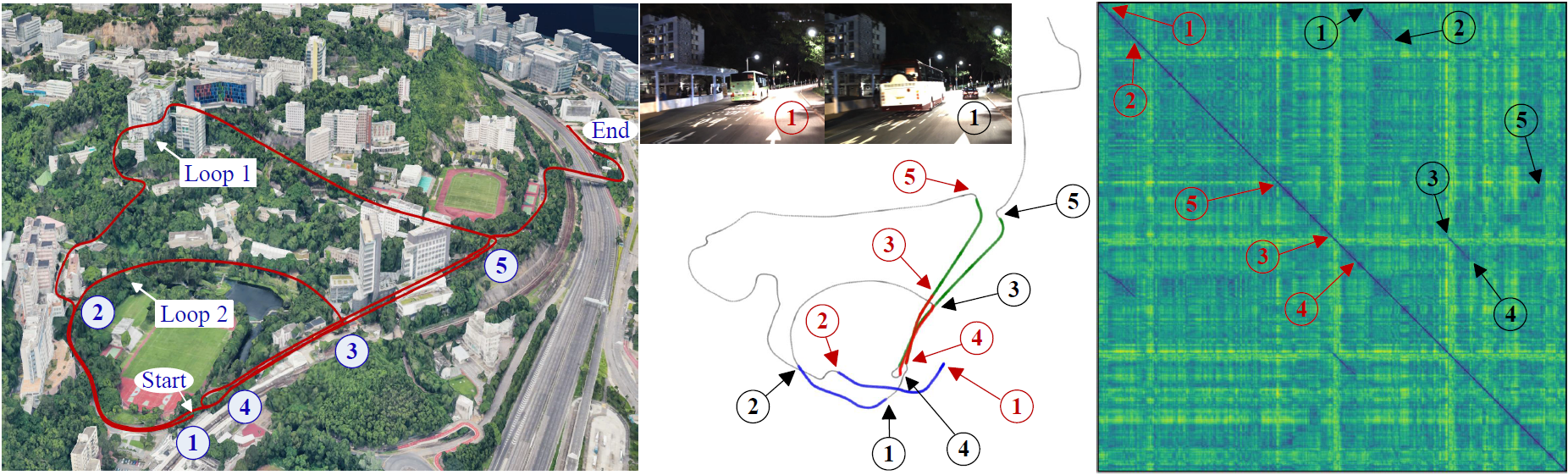}
\caption{Experiment results in the university campus. Left: The campus environment in Google Map and the vehicle route. Middle: The trajectory recorded in the point cloud map built by LeGO-LOAM \cite{legoloam} and two video frames recorded at the same place in the different experiment loops, where the red markers belong to the first loop and the black markers belong to the second loop. Right: The $L_2$ distances between the generated global descriptors along the whole point cloud frame sequence.}
\label{figexp2}
\vspace{-0.3cm}
\end{figure*}

\begin{figure}[!h]
	\centering
	\includegraphics[width=1\columnwidth]{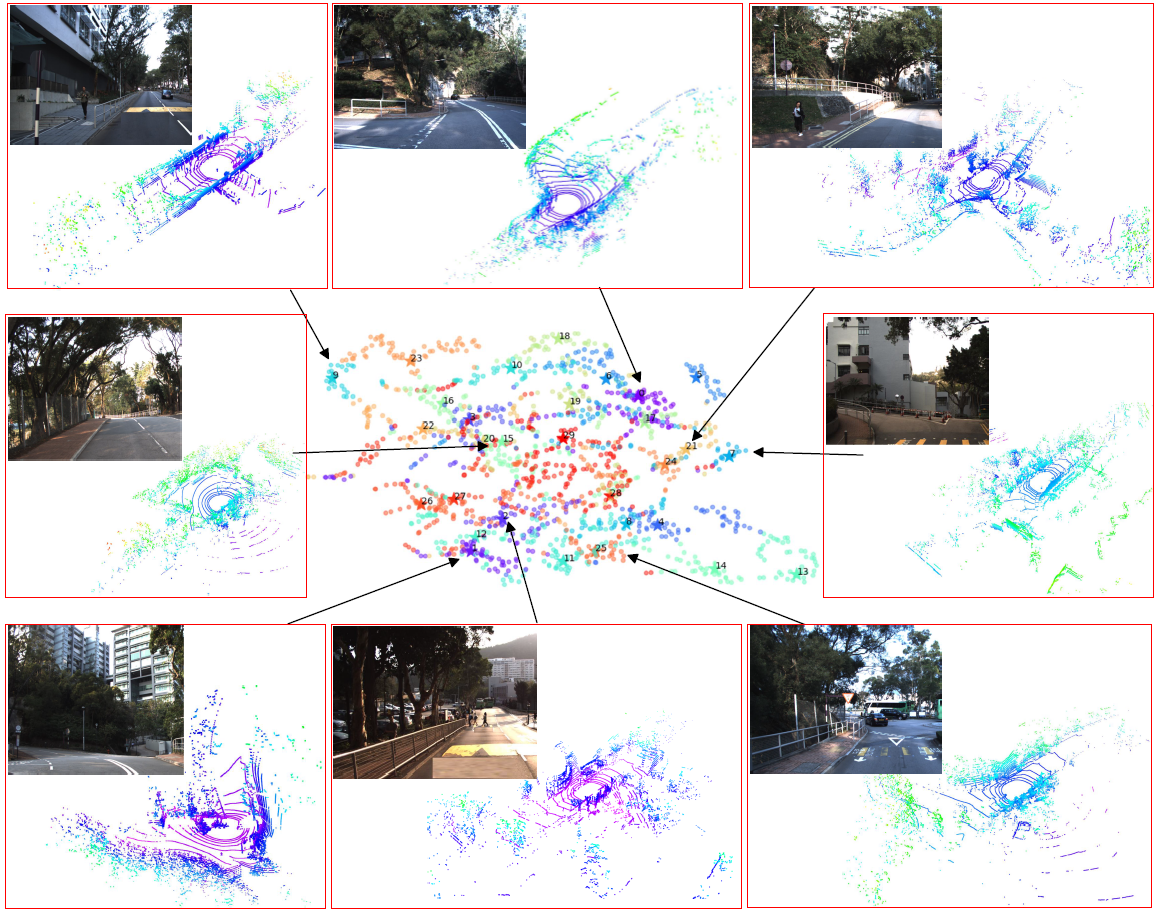}
	\caption{The visualization of the selected typical places in the experiments. Middle figure shows the t-SNE visualization of the clustering results using the generated global descriptors. The color of each point represents its belonged cluster. The eight point clouds around are typical places which correspond to the largest eight clusters (there are 29 clusters in total), we also show their corresponding images.}
	\label{figclu2}
\end{figure}

\begin{figure}[!h]
\centering
\includegraphics[width=1\columnwidth]{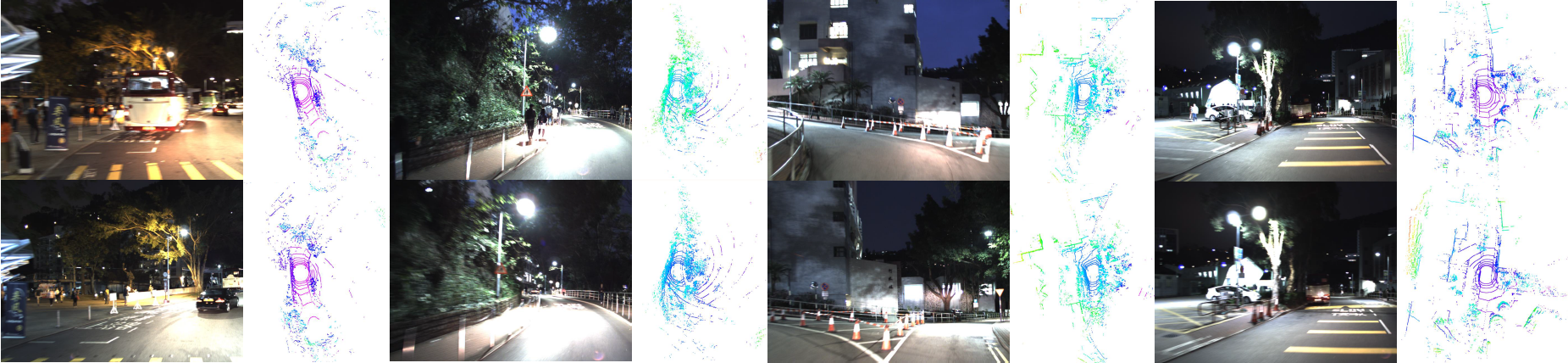}
\caption{An example of the matched sequences in our loop-closure detection results, where the video image frames and the corresponding point cloud frames recorded in the first experiment loop are shown in the Upper, while the frames recorded in the second loop are shown in the Lower. All the point clouds have been projected into the horizontal plane for better visualization.}
\label{figexp3}
\vspace{-0.6cm}
\end{figure}

We conduct experiments in the campus of The Chinese University of Hong Kong to validate the proposed approach in large-scale outdoor environments. The desired vehicle velocity is set to $V_d=10m/s$. In experiments, we use the loop-closure detection approach presented in Section V, where the trajectory velocity bounds in sequence searching are set to $V_{min}=0.8V_d$ and $V_{max}=1.2V_d$, and the sequence searching window size is set to 10 point cloud frames. In our platform, the total computation time of local feature extraction, global descriptor generation and coarse-to-fine sequence matching is about 500ms. Note that the university campus is built on the mountain area, the slope terrain brings large difficulties in the loop-closure tasks. The map built by the state-of-the-art LeGO-LOAM \cite{legoloam} is shown in the middle of Fig. \ref{figexp2}, we can find that the map can not be loop-closed due to the slope terrain and large distances. However, in the right of Fig. \ref{figexp2}, the proposed approach finds out three matching segments successfully, and in particular, the same route segment with the opposite moving directions can also be recognized (as shown in the segment from location 5 to location 4 in the first loop and in the segment from location 4 to location 5 in the second loop), these validate the effectiveness and robustness of the proposed approach, and will greatly facilitate the practical applications. The detailed place clustering and typical place selection results are shown in Fig. \ref{figclu2}. Fig. \ref{figexp3} gives an example to show the corresponding point cloud frames and video image frames from a matched sequence.

\section{Discussion and Conclusion}

In this paper, we propose a loop-closure detection solution based on 3D point cloud learning and coarse-to-fine sequence matching. 
The dataset testing and experiment results demonstrate that our approach has advanced performances for point cloud-based place recognition and loop-closure detection in the large-scale environment, exceeding PointNetVLAD and LeGO-LOAM. Also please note that we do not use any global position information or odometry data in the loop-closure detection. These validate the effectiveness and practical applicability of the proposed approach.
What's more, the model trained on the RobotCar dataset can be directly applied to the real-world applications without any further training, which greatly facilitates the practical applications.
In the future, we will implement the proposed approach to solve the dynamic and life-long SLAM problems.


\end{document}